\pdfoutput=1

\documentclass[11pt]{article}

\usepackage[final]{acl}

\usepackage{times}
\usepackage{latexsym}

\usepackage[T1]{fontenc}

\usepackage[utf8]{inputenc}

\usepackage{microtype}

\usepackage{inconsolata}

\usepackage{graphicx}
\DeclareUnicodeCharacter{2060}{\nolinebreak}
\usepackage{multirow} 

\title{Asking Again and Again:\\Exploring LLM Robustness to Repeated Questions}

\author{Sagi Shaier$^\nabla$, Mario Sanz-Guerrero$^{\diamondsuit}$, Katharina von der Wense$^{\nabla\diamondsuit}$ \\
  $^\nabla$University of Colorado Boulder\\
$^\diamondsuit$Johannes Gutenberg University Mainz\\
$^\nabla$E-mail: \{sagi.shaier, katharina.kann\}@colorado.edu \\
$^{\diamondsuit}$E-mail: msanzgue@uni-mainz.de
 \\}

\begin{document}
\maketitle
\begin{abstract}
This study investigates whether repeating questions within prompts influences the performance of large language models (LLMs). We hypothesize that reiterating a question within a single prompt might enhance the model's focus on key elements of the query. We evaluate five recent LLMs---including GPT-4o-mini, DeepSeek-V3, and smaller open-source models---on three reading comprehension datasets under different prompt settings, varying question repetition levels (1, 3, or 5 times per prompt). Our results demonstrate that question repetition can increase models' accuracy by up to $6\%$. However, across all models, settings, and datasets, we do not find the result statistically significant. These findings provide insights into prompt design and LLM behavior, suggesting that repetition alone does not significantly impact output quality. 
\end{abstract}

\section{Introduction}
Large language models (LLMs) have become indispensable tools across various fields, excelling in tasks such as natural language understanding \cite{longchat2023, openai2023, openai2023gpt4, anthropic}, content creation \cite{ma-etal-2024-mops, bae-kim-2024-collective}, and question answering \cite{openai2023, openai2023gpt4}. Their capability to generate coherent, human-like responses has made them particularly valuable in applications ranging from chatbots to research support. Among these applications, question answering stands out as a key area, highlighting the models' strengths in reasoning, information retrieval, and contextual comprehension.

\begin{figure}[t]
    \includegraphics[width=\columnwidth]{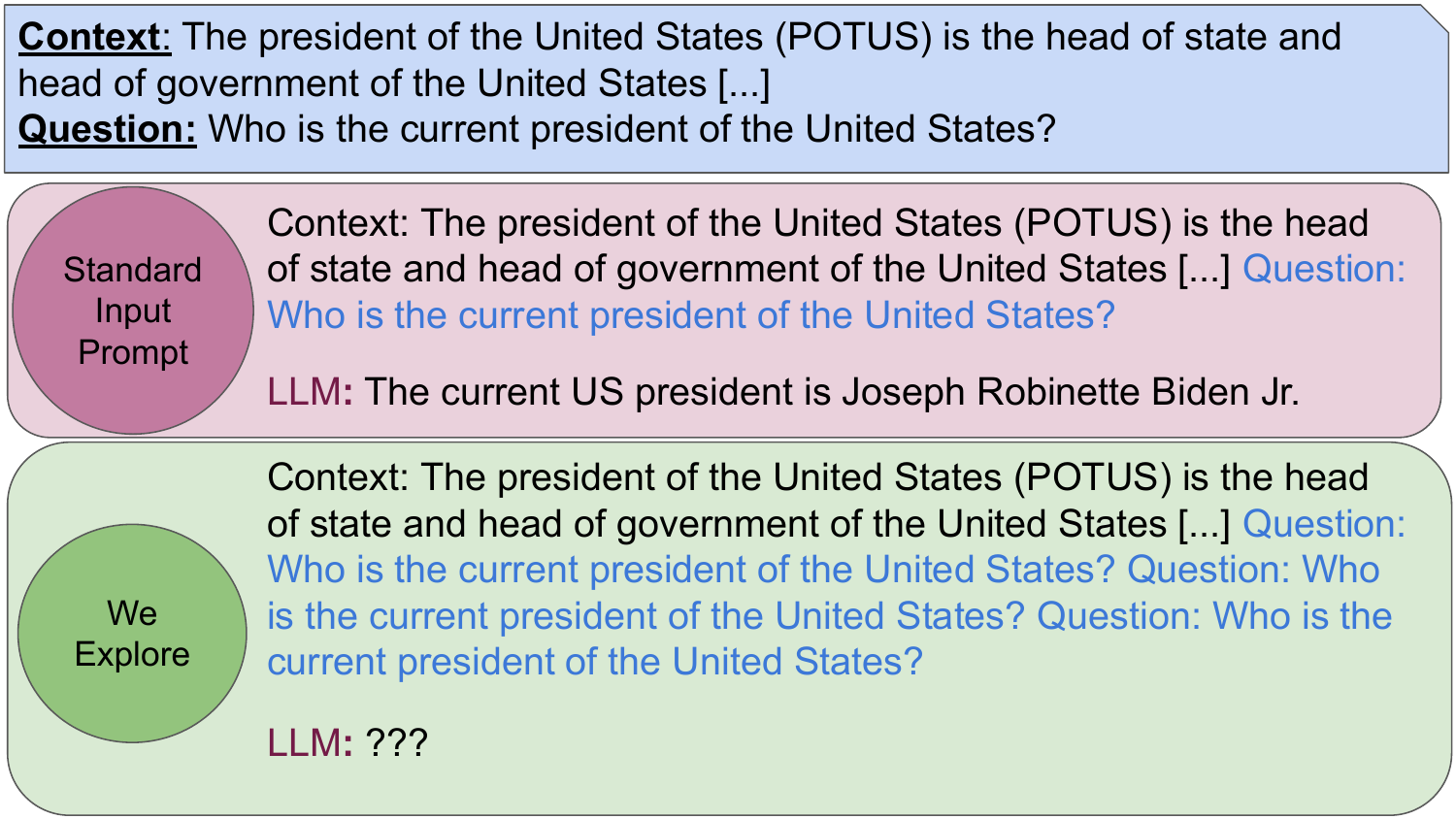}
    \caption{Overview of our study on question repetition in LLM prompts. We evaluate five recent LLMs across three datasets and four prompt settings, varying question repetition (1, 3, or 5 times) to assess its impact on performance.}
    \label{main_figure}
\end{figure}

Interactions with LLMs typically involve providing context and posing questions. Prior research has shown that LLMs are sensitive to input variations, such as the order of questions and context \cite{shaier-etal-2024-say}, conflicting information \cite{longpre-etal-2021-entity, zhou2023contextfaithful, dist, shaier-etal-2024-desiderata, shaier-etal-2024-adaptive, hong2023discern, chen-etal-2022-rich}, and minor adversarial perturbations \cite{jia-liang-2017-adversarial, cao2022tasa, 10066742}. These variations can lead to unstable behavior, such as degraded performance or the introduction of biases \cite{levy-etal-2024-evaluating, shaier-etal-2023-emerging}. This body of work highlights the importance of input structure and presentation in shaping response quality and relevance. However, the specific effect of question repetition within prompts remains under-explored.

This study investigates whether repeating a question within a prompt can improve LLM performance. By systematically increasing the frequency of question repetition, we evaluate its impact on overall performance. Specifically, we test five recent LLMs on three reading comprehension datasets under four different settings---including open- and closed-book---, varying the repetition of each question to 1, 3, or 5 times per prompt.

Our results demonstrate that repeating questions within a single prompt, while increasing some models' accuracy by up to $6\%$, does not improve model performance significantly, underscoring the stability of the tested models. This finding suggests that LLMs are not easily influenced by repeated input structures and that repeating questions does not encourage them to ``focus more'' on the repeated information. This contrasts with prior research \cite{mekala2024echopromptinstructingmodelrephrase}, which emphasizes the benefits of instructing models to restate questions in their responses. These insights highlight the need for further research to explore the nuanced relationship between question repetition and LLM behavior.


\section{Experiments}
In our experiments, each dataset consists of triples $(q, c, a)$, where $q$ is a question, $c$ is a context document, and $a$ is the gold answer. Following prior work \citep{brown2020fewshot, chowdhery2022palm}, we concatenate the question and context into a single string as input to the model.

The primary objective of this study is to evaluate the impact of question repetition on LLM performance. Specifically, we experiment with repeating the question 1, 3, and 5 times to assess its effect.

\subsection{Prompt Configurations}
\label{subsec:configurations}
We test four prompt configurations to evaluate LLM performance under varying question repetition strategies:

\paragraph{Open-Book}
In the open-book setting, the model is provided with both the question and its original dataset context. Consistent with prior findings \cite{shaier-etal-2024-say}, we place the context before the question, as this ordering has been shown to improve performance. The prompt structure is detailed in Appendix \ref{app:open_book_prompt}.

\paragraph{Closed-Book}
In the closed-book configuration, the model relies solely on its internal knowledge, as no context is provided. The prompt format is outlined in Appendix \ref{app:closed_book_prompt}.

\paragraph{Question-Context-Question (QCQ)}
This setting extends the open-book approach by presenting the question both before and after the context. This allows us to analyze whether repetition in different parts of the prompt enhances performance. The prompt structure is detailed in Appendix \ref{app:qcq_prompt}.

\paragraph{Paraphrasing}
In this configuration, we first generate paraphrased versions of the original question using the model itself. These paraphrased questions, along with the original question, are then appended to the context in the prompt, similar to the open-book setting. This setting aims to assess whether paraphrasing the question can improve the model's understanding and response quality. The prompt structure is provided in Appendix \ref{app:paraphrasing_prompt}.

\subsection{Metrics}
\label{subsec:metrics}
To measure accuracy, we adopt substring matching, as used in prior work \citep{liu2023lost, mallen-etal-2023-trust, kandpal2023large}. This method checks whether any of the gold answers appear in the model’s output.

\subsection{Datasets}
\label{subsec:datasets}
We evaluate our models on three datasets: Stanford Question Answering Dataset (SQuAD) \cite{rajpurkar-etal-2016-squad}, HotPotQA \cite{yang-etal-2018-hotpotqa}, and Natural Questions (NQ) \cite{kwiatkowski-etal-2019-natural}. Due to the high cost of using LLMs and limited funding, we sample 500 questions from each dataset. With 180 experimental settings---3 levels of question repetition × 4 configurations (\S \ref{subsec:configurations}) × 3 datasets (\S \ref{subsec:datasets}) × 5 models (\S \ref{subsec:models})---this setup results in a total of 90,000 questions. Further details about the datasets are provided in Appendix \ref{app:datasets}.

\subsection{Models}
\label{subsec:models}
Our experiments involve five models, spanning large-scale and smaller open-source models. For large-scale models, we use GPT-4o-mini\footnote{Specific checkpoint: \texttt{gpt-4o-mini-2024-07-18}} \cite{openai2023gpt4}, a recent ChatGPT version accessed via the OpenAI API, achieving $82\%$ on MMLU and excelling in chat preference ratings; and DeepSeek-V3 \cite{deepseekai2024deepseekv3}, a 685B parameter model with $88.5\%$ on MMLU that surpasses GPT-4o in various tasks. These models provide a robust baseline for large-scale performance.

For smaller open-source models, we include Llama-3.1 (8B) \cite{dubey2024llama3}, Mistral 7B \cite{jiang2023mistral7b}, and Phi-4 (14B) \cite{abdin2024phi4}, with Phi-4 achieving $85\%$ on MMLU and outperforming GPT-4o on STEM QA tasks. Open-source models were run locally on an NVIDIA A100 GPU. We analyze how model size affects sensitivity to input patterns under varying levels of repetition.

\section{Results}

\begin{table*}
    \small
    \centering
    \begin{tabular}{|c|c||c|c|c||c|c|c||c|c|c|}
    \hline
    \multirow{2}{*}{\textbf{Model}} & \multirow{2}{*}{\textbf{Configuration}} & \multicolumn{3}{c||}{\textbf{HotPotQA}} & \multicolumn{3}{c||}{\textbf{SQuAD}} & \multicolumn{3}{c|}{\textbf{Natural Questions}} \\ \cline{3-11}
    & & Qx1 & Qx3 & Qx5 & Qx1 & Qx3 & Qx5 & Qx1 & Qx3 & Qx5 \\ \hline\hline
    \multirow{4}{*}{GPT-4o-mini} 
    & Open-Book & 0.58 & 0.58 & 0.59 & 0.99 & 0.99 & 0.98 & 0.68 & 0.69 & 0.70 \\
    & Closed-Book & 0.42 & 0.42 & 0.43 & 0.49 & 0.49 & 0.49 & 0.42 & 0.41 & 0.42 \\
    & QCQ & 0.56 & 0.56 & 0.57 & 0.96 & 0.95 & 0.96 & 0.69 & 0.69 & 0.70 \\
    & Paraphrasing & 0.59 & 0.57 & 0.57 & 0.97 & 0.98 & 0.97 & 0.69 & 0.67 & 0.68 \\
    \hline
    \multirow{4}{*}{DeepSeek-V3} 
    & Open-Book & 0.63 & 0.63 & 0.62 & 0.98 & 0.97 & 0.97 & 0.73 & 0.72 & 0.72 \\
    & Closed-Book & 0.50 & 0.49 & 0.49 & 0.55 & 0.55 & 0.55 & 0.47 & 0.47 & 0.46 \\
    & QCQ & 0.63 & 0.61 & 0.62 & 0.97 & 0.97 & 0.96 & 0.72 & 0.71 & 0.71 \\
    & Paraphrasing & 0.63 & 0.61 & 0.63 & 0.98 & 0.95 & 0.95 & 0.73 & 0.70 & 0.69 \\
    \hline
    \multirow{4}{*}{Llama-3.1} 
    & Open-Book & 0.52 & 0.53 & 0.52 & 0.90 & 0.92 & 0.88 & 0.67 & 0.70 & 0.70 \\
    & Closed-Book & 0.24 & 0.28 & 0.30 & 0.31 & 0.33 & 0.33 & 0.38 & 0.41 & 0.41 \\
    & QCQ & 0.53 & 0.53 & 0.53 & 0.90 & 0.90 & 0.91 & 0.69 & 0.71 & 0.71 \\
    & Paraphrasing & 0.52 & 0.55 & 0.52 & 0.91 & 0.92 & 0.91 & 0.67 & 0.68 & 0.69 \\
    \hline
    \multirow{4}{*}{Mistral 7B} 
    & Open-Book & 0.51 & 0.48 & 0.48 & 0.94 & 0.92 & 0.93 & 0.68 & 0.68 & 0.67 \\
    & Closed-Book & 0.34 & 0.35 & 0.34 & 0.42 & 0.43 & 0.41 & 0.34 & 0.33 & 0.33 \\
    & QCQ & 0.51 & 0.51 & 0.50 & 0.91 & 0.90 & 0.91 & 0.67 & 0.65 & 0.64 \\
    & Paraphrasing & 0.51 & 0.48 & 0.48 & 0.93 & 0.93 & 0.93 & 0.68 & 0.66 & 0.67 \\
    \hline
    \multirow{4}{*}{Phi-4} 
    & Open-Book & 0.57 & 0.58 & 0.58 & 0.98 & 0.98 & 0.98 & 0.70 & 0.70 & 0.71 \\
    & Closed-Book & 0.34 & 0.35 & 0.36 & 0.37 & 0.40 & 0.41 & 0.35 & 0.34 & 0.34 \\
    & QCQ & 0.57 & 0.57 & 0.57 & 0.97 & 0.97 & 0.98 & 0.68 & 0.69 & 0.70 \\
    & Paraphrasing & 0.57 & 0.58 & 0.58 & 0.98 & 0.97 & 0.96 & 0.70 & 0.70 & 0.69 \\
    \hline
    \end{tabular}
    \caption{Accuracy of various LLMs on three datasets with varying levels of question repetition (Qx1, Qx3, Qx5) and prompt configurations.}
    \label{tab:results}
\end{table*}

\begin{figure}
    \includegraphics[width=\columnwidth]{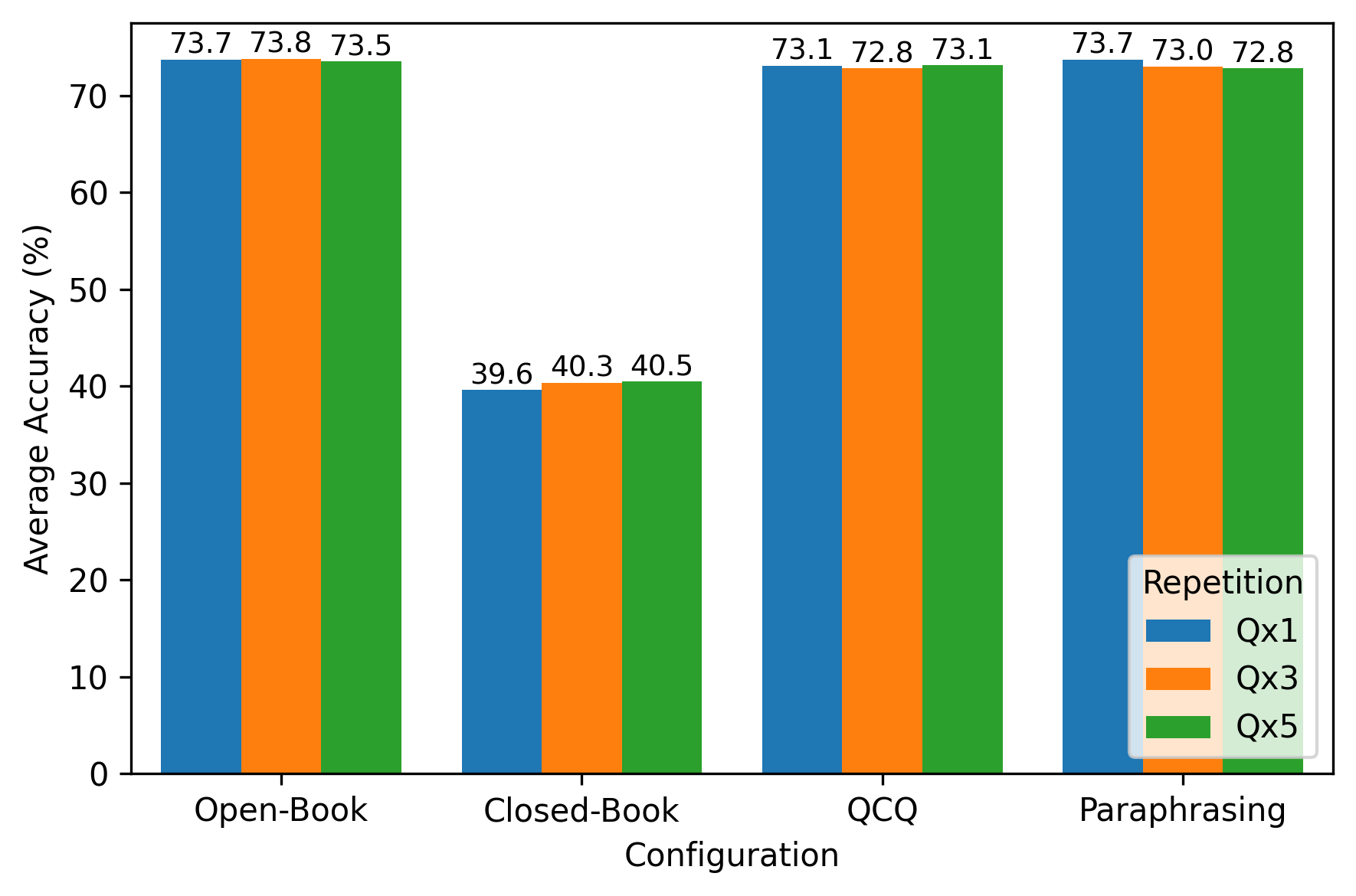}
    \caption{Average accuracy by configuration and repetition level across all models and datasets.}
    \label{fig:accuracy_by_configuration}
\end{figure}

This section presents the results of our experiments assessing the impact of repeated question exposure on LLM performance across different prompt configurations and datasets. Table \ref{tab:results} provides the accuracy results of repetition levels (Qx1, Qx3, Qx5) and configurations for each dataset and model, and Figure \ref{fig:accuracy_by_configuration} summarizes the findings.

\subsection{Comparison Across Configurations}

Our results show that repeating questions have no significant effect on LLM performance across different settings.

In the open-book setting, performance remain stable across repetition levels for all datasets and models, with only minor variations within the margin of error. Smaller models show more fluctuation, though not consistently positive. For instance, Llama-3.1's accuracy increases by 3\% on NQ with three or five repetitions, while Mistral's accuracy decrease by 3\% on HotPotQA under the same conditions. These findings highlight the robustness of LLMs to repeated questions.

The closed-book setting show the lowest performance overall, as models rely solely on internal knowledge. This setting exhibit the largest (though still minor) improvements with repetition, such as a 6\% gain for Llama-3.1 on HotPotQA and a 4\% gain for Phi-4 on SQuAD with five repetitions. Larger models remained less sensitive to repetition, and differences for smaller models were not significant. These results suggest that repetition has no meaningful impact on closed-book performance, reinforcing that it neither improves nor harms output quality.

In the QCQ setting, where the question is repeated before and after the context, performances remain stable across repetition levels. No significant improvements were observed, and the most notable change is a 3\% decrease for Mistral 7B on NQ with five repetitions. Contrary to expectations, QCQ did not outperform the open-book setting, where performance is nearly identical across both.

Finally, the paraphrasing setting, which provides paraphrased versions of the original question, show no significant performance gains. Interestingly, this is the only setting where larger models exhibit noticeable differences with repetition. For example, GPT-4o-mini's accuracy decrease by 2\% on HotPotQA, and DeepSeek-V3's decrease by 3\% on NQ with five repetitions. This suggests that paraphrasing may introduce noise, negatively affecting performance, contrary to our hypothesis that it would enhance understanding and response quality.

\subsection{Differences Between Datasets and Models}

\begin{figure}
    \includegraphics[width=\columnwidth]{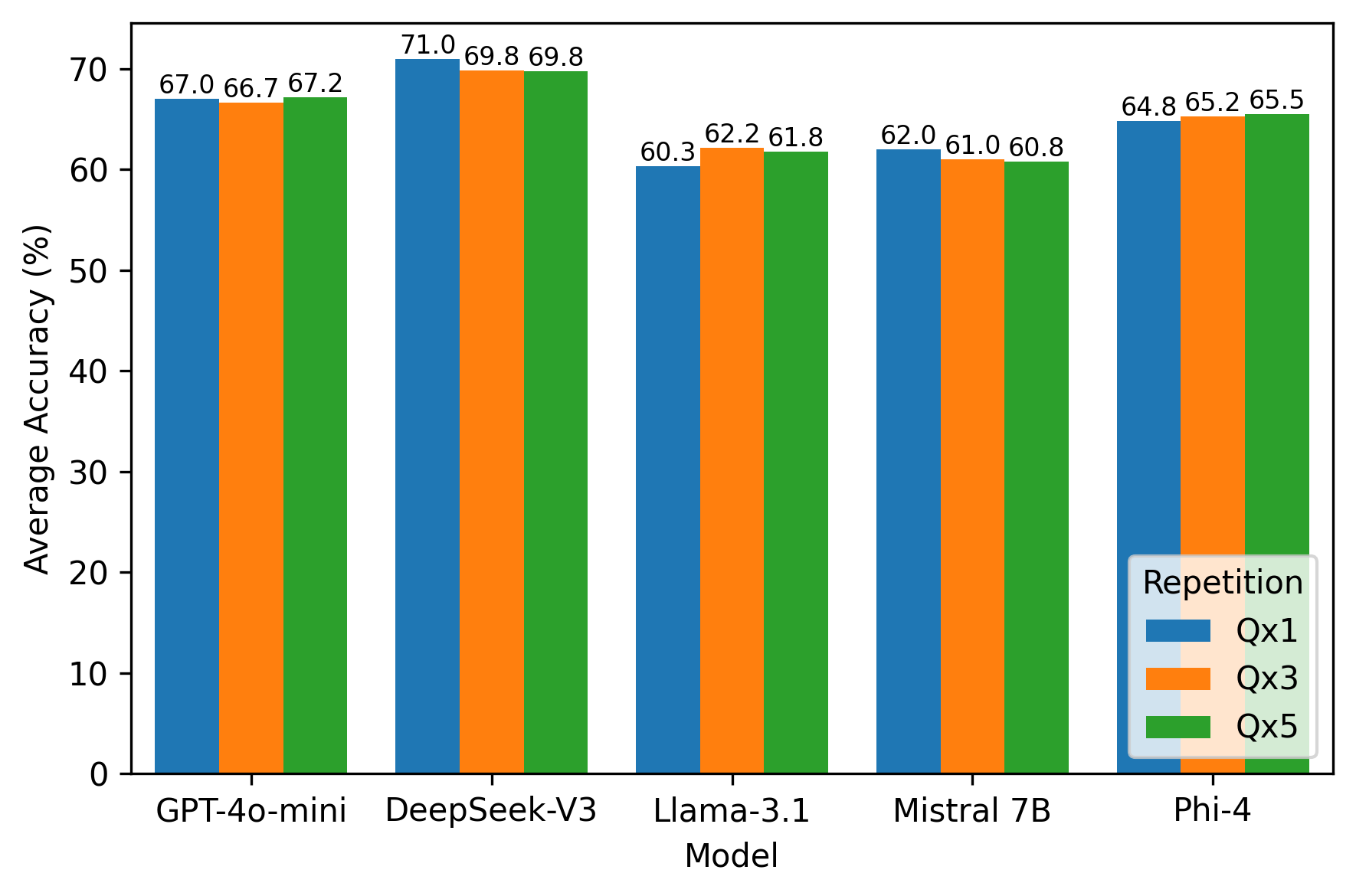}
    \caption{Average accuracy by model and repetition level across all configurations and datasets.}
    \label{fig:accuracy_by_model}
\end{figure}

As shown in Figure \ref{fig:accuracy_by_model}, DeepSeek-V3 consistently outperforms other models across all datasets and repetition levels. The largest performance gap occurs in the closed-book setting, highlighting its stronger internal representation of information and ability to handle questions without context. In contrast, smaller models like Llama-3.1 struggle in this setting, likely due to their limited size and parametric knowledge. Notably, Phi-4 performs nearly as well as DeepSeek-V3 and surpasses GPT-4o-mini in settings with context, suggesting diminishing returns for scaling model size on these tasks. Further analysis reveals the most significant performance gaps between large and small models in the closed-book setting (Appendix \ref{app:extended_results}).

Models achieve higher accuracy on SQuAD than on HotPotQA and NQ in most configurations. This reflects the greater complexity of HotPotQA, which requires multi-hop reasoning,
compared to SQuAD's single-passage questions. NQ show the smallest performance differences across models (see Appendix \ref{app:extended_results}), with an apparent upper bound of around 70\% accuracy for both large and small LLMs. For example, Llama-3.1, the worst-performing model on average, matches GPT-4o-mini's performance on NQ. This suggests that the dataset's complexity or design limits further gains, and context enables smaller models to perform nearly as well as larger ones, reinforcing the trend of diminishing returns with model scaling.

\subsection{Summary of Findings}
Our experiments demonstrate that question repetition within a prompt neither significantly improves nor significantly  degrades the performance of LLMs across all models, datasets, and settings. The consistent accuracies observed across Qx1, Qx3, and Qx5 configurations suggest that LLMs process information effectively without being influenced by repeated phrasing. Statistical tests (see Appendix \ref{app:statistical_tests} for details) confirm that the differences between repeating questions 1, 3, or 5 times were not significant, further supporting this conclusion. This stability highlights their robustness to redundant input structures, contrasting with prior work \cite{mekala2024echopromptinstructingmodelrephrase}, which emphasizes the benefits of instructing models to restate questions in their responses.

Overall, our findings indicate that question repetition has no meaningful impact on performance, regardless of dataset complexity, model size, context availability, or repetition level. These results underscore the stability of LLMs in handling repeated and paraphrased input structures, providing valuable insights into their interaction with variations in prompt design.


\section{Conclusion}
In this study, we explore the impact of question repetition within prompts on the performance of recent LLMs using the HotPotQA, SQuAD, and Natural Questions datasets across different prompt configurations. Our results demonstrate that repeating questions neither improves nor degrades performance, highlighting the robustness of LLMs to redundant input structures. This finding contrasts with prior work that emphasizes the benefits of instructing models to restate questions in their responses, suggesting that repetition alone does not influence output quality. These insights contribute to a deeper understanding of how LLMs interact with prompt design, offering a foundation for optimizing question-answering systems.

Future work can explore the effects of question repetition on LLMs with different architectures or training paradigms, as well as expand the evaluation to include more diverse datasets, such as those with open-ended or subjective answers. Additionally, a detailed investigation into the interplay between prompt structure, repetition, and interpretability can provide actionable insights for tailoring prompts to specific applications. By addressing these directions, researchers can further refine our understanding of LLM behavior and enhance their practical utility.

\section*{Limitations}
While this work offers insights into question repetition effects, several limitations warrant consideration. First,
the study focuses exclusively on reading comprehension tasks. However, repetition effects might differ in reasoning-heavy or creative domains. Second, while we evaluate recent models, the rapid evolution of LLM architectures means newer systems could exhibit altered sensitivity to repetition patterns. Lastly, we focus on causal LLMs, primarily because a significant portion of the field is interested in them. However, the effect of question repetition remains unexplored in masked LMs.


\bibliography{custom}

\appendix

\section{Prompt Structures}
\label{app:prompt_structures}

\subsection{Open-Book Prompt}
\label{app:open_book_prompt}

\texttt{\noindent
    \textbf{Context}: \{context\} \\
    (\textbf{Question}: \{question\}) $\times$ repetitions
}

\subsubsection*{Example (Natural Questions, Qx3 setting)\footnote{The selected example provides a concise and straightforward context, specifically chosen to facilitate illustration.}:}
\texttt{\noindent
    \underline{Context}: Deadpool 2 is scheduled to be released in the United States on May 18, 2018. A sequel, Deadpool 3, is in development. \underline{Question}: When is the next Deadpool movie being released? \underline{Question}: When is the next Deadpool movie being released? \underline{Question}: When is the next Deadpool movie being released?
}

\subsection{Closed-Book Prompt}
\label{app:closed_book_prompt}

\texttt{\noindent
    (\textbf{Question}: \{question\}) $\times$ repetitions
}

\subsubsection*{Example (Natural Questions, Qx3 setting):}
\texttt{\noindent
    \underline{Question}: When is the next Deadpool movie being released? \underline{Question}: When is the next Deadpool movie being released? \underline{Question}: When is the next Deadpool movie being released?
}

\subsection{Question-Context-Question (QCQ) Prompt}
\label{app:qcq_prompt}

\texttt{\noindent
    (\textbf{Question}: \{question\}) $\times$ repetitions \\
    \textbf{Context}: \{context\} \\
    (\textbf{Question}: \{question\}) $\times$ repetitions
}

\subsubsection*{Example (Natural Questions, Qx3 setting):}
\texttt{\noindent
    \underline{Question}: When is the next Deadpool movie being released? \underline{Question}: When is the next Deadpool movie being released? \underline{Question}: When is the next Deadpool movie being released? \underline{Context}: Deadpool 2 is scheduled to be released in the United States on May 18, 2018. A sequel, Deadpool 3, is in development. \underline{Question}: When is the next Deadpool movie being released? \underline{Question}: When is the next Deadpool movie being released? \underline{Question}: When is the next Deadpool movie being released?
}

\subsection{Paraphrasing Prompt}
\label{app:paraphrasing_prompt}

\texttt{\noindent
    \textbf{Context}: \{context\} \\
    \textbf{Question}: \{original\_question\} \\(\textbf{Question}: \{paraphrased\_question[i]\}) $\times$ repetitions
}

\subsubsection*{Example (Natural Questions, Qx3 setting):}
\texttt{\noindent
    \underline{Context}: Deadpool 2 is scheduled to be released in the United States on May 18, 2018. A sequel, Deadpool 3, is in development. \underline{Question}: When is the next Deadpool movie being released? \underline{Question}: What is the release date for the next Deadpool movie? \underline{Question}: When can we expect the next Deadpool film to be released?
}

\section{Dataset Details}
\label{app:datasets}
\paragraph{Stanford Question Answering Dataset (SQuAD)}
SQuAD \cite{rajpurkar-etal-2016-squad} is a large-scale reading comprehension dataset containing over 100,000 questions crafted by crowdworkers based on a diverse set of Wikipedia articles. Each question is designed such that the answer is a specific text segment from the corresponding article, encouraging models to closely analyze and comprehend the context of the passage. The dataset includes a wide range of question types, requiring reasoning at various levels of complexity, from simple fact retrieval to more nuanced understanding of dependencies within the text.

\paragraph{HotPotQA}
HotPotQA \cite{yang-etal-2018-hotpotqa} is a complex multi-hop question-answering dataset comprising 113,000 question-answer pairs grounded in Wikipedia articles. It is specifically designed to address limitations in existing question answering datasets by emphasizing multi-hop reasoning, where answering a question requires synthesizing information from multiple supporting documents. The dataset stands out for its diversity, as it is not restricted to predefined knowledge bases or schemas, ensuring that questions reflect realistic and open-ended scenarios. Additionally, HotPotQA provides annotated sentence-level supporting facts, enabling strong supervision for reasoning and facilitating explainable predictions.

\paragraph{Natural Questions (NQ)}  
The Natural Questions dataset \cite{kwiatkowski-etal-2019-natural} consists of real, anonymized queries submitted to the Google search engine, paired with full Wikipedia pages and annotated long and short answers. Given the extensive length of contexts in NQ and the associated computational expense, we follow prior work \cite{shaier-etal-2024-say} by using the long answers as context for each question and the corresponding short answers as the gold labels. This dataset is a standard benchmark widely used in the research community, enabling reproducibility and facilitating comparisons with prior work. Its diverse range of question types and realistic contexts makes it an excellent resource for evaluating the generalizability of our approach.

\section{Extended Results}
\label{app:extended_results}

\begin{figure}
    \includegraphics[width=\columnwidth]{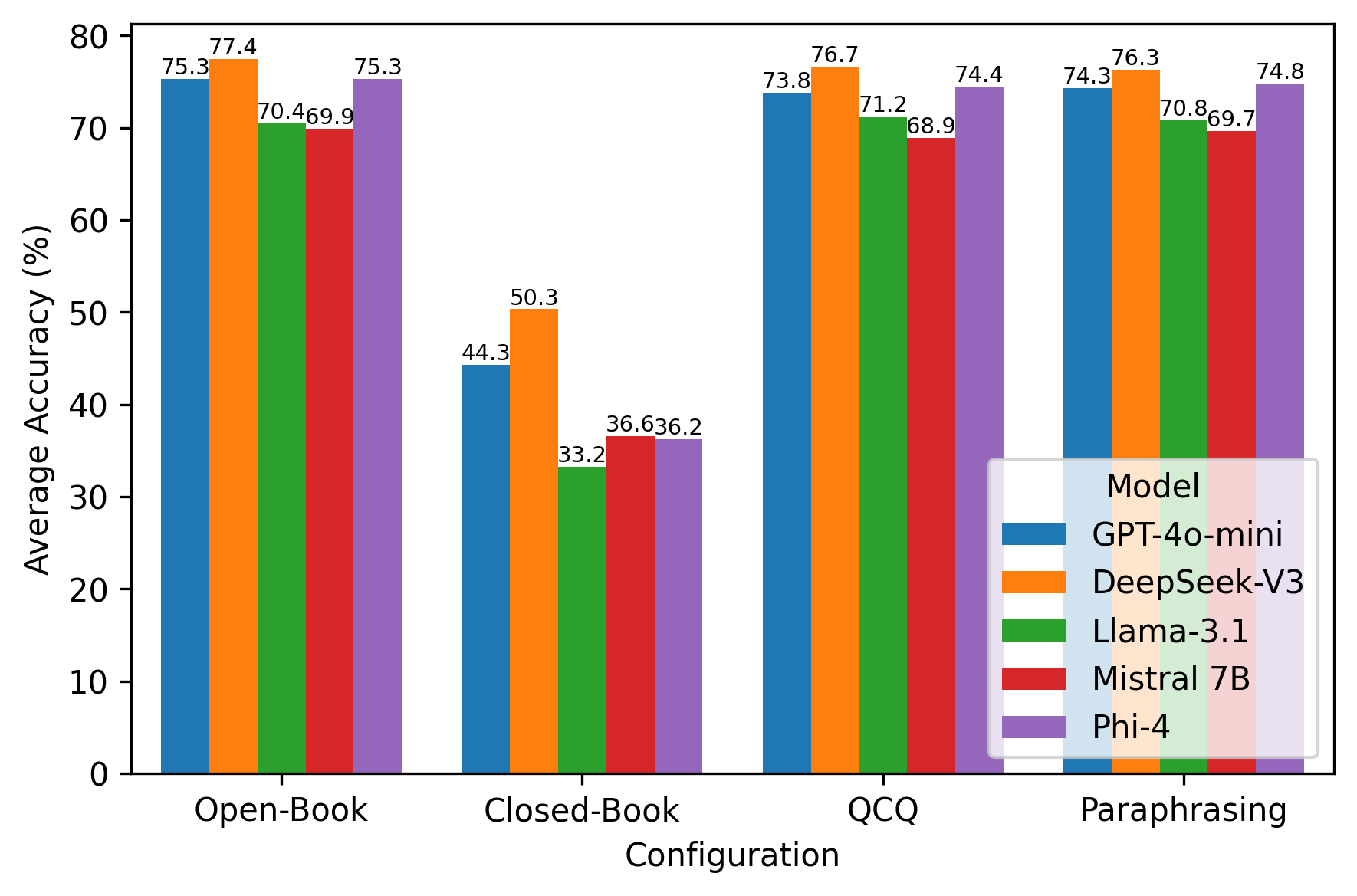}
    \caption{Average accuracy by configuration and model across all datasets and repetition levels.}
    \label{fig:accuracy_by_configuration_and_model}
\end{figure}

Figure \ref{fig:accuracy_by_configuration_and_model} displays the average accuracy of each model, grouped by configuration, across all datasets and repetition levels. The results reveal that the closed-book setting consistently yielded the lowest performance, as models relied solely on their internal knowledge. This setting also exhibited the largest performance gaps between large and small models, aligning with the expectation that larger models possess a stronger internal representation of information and are better equipped to handle questions without context. In contrast, smaller models performed comparably to larger ones in other settings, particularly Phi-4, which nearly matched DeepSeek-V3 and even surpassed GPT-4o-mini. This suggests diminishing returns for scaling model size on these tasks.

\begin{figure}
    \includegraphics[width=\columnwidth]{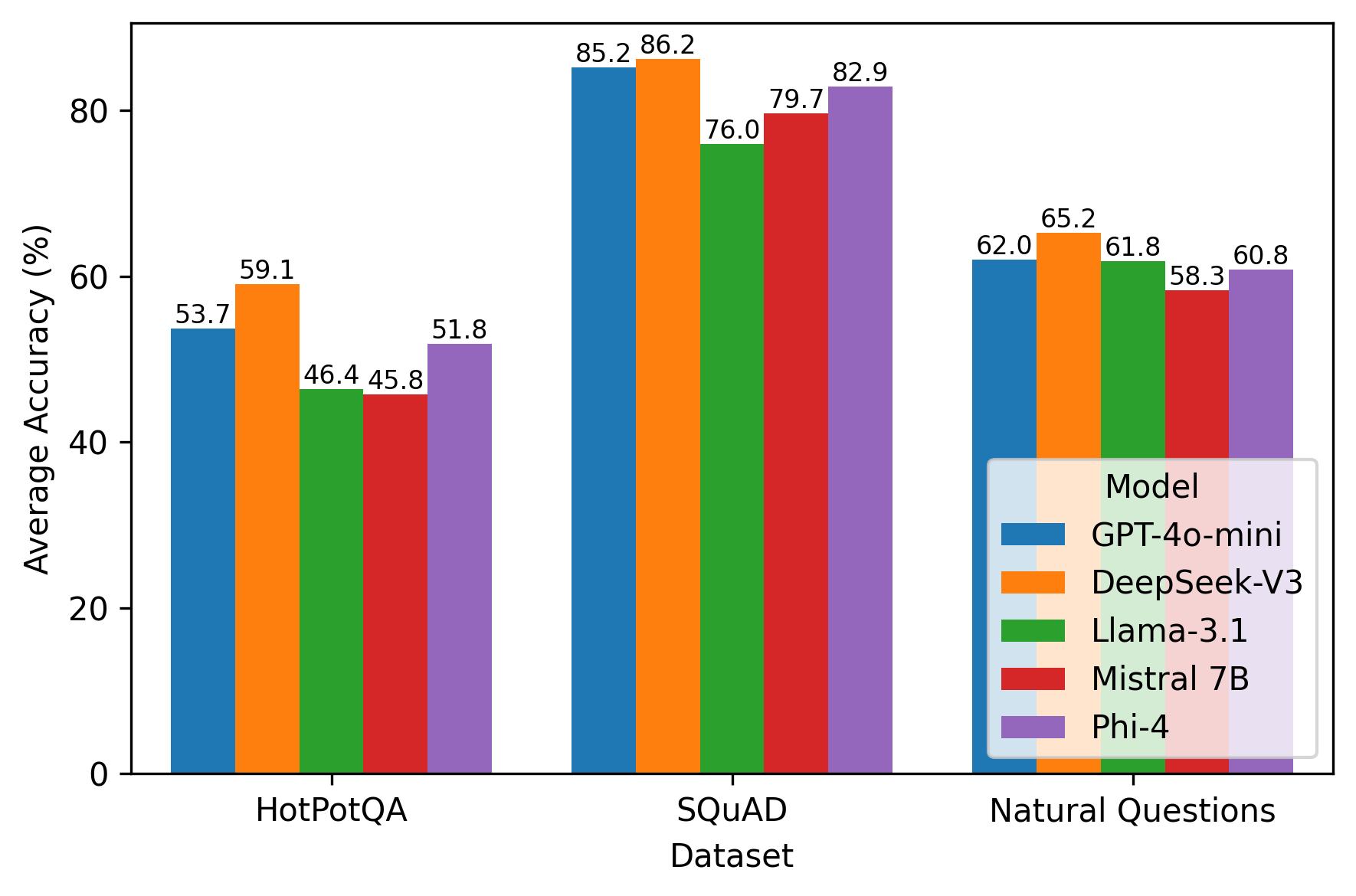}
    \caption{Average accuracy by dataset and model across all configurations and repetition levels.}
    \label{fig:accuracy_by_dataset_and_model}
\end{figure}

Figure \ref{fig:accuracy_by_dataset_and_model} presents the average accuracy of each model, grouped by dataset, across all configurations and repetition levels. Models achieved higher accuracy on SQuAD than on HotPotQA and NQ in most configurations, reflecting the greater complexity of HotPotQA, which requires multi-hop reasoning across multiple documents, compared to SQuAD's single-passage questions. NQ showed the smallest performance differences across models, with an apparent upper bound of around 70\% accuracy for both large and small LLMs. For instance, Llama-3.1, the worst-performing model on average, matched GPT-4o-mini's performance on NQ. This suggests that the dataset's complexity or design may limit further gains, and context enables smaller models to perform nearly as well as larger ones, reinforcing the trend of diminishing returns with model scaling.

\subsection{Statistical Tests}
\label{app:statistical_tests}

To assess the significance of differences in model performance across question repetition levels (Qx1, Qx3, and Qx5), we conducted statistical tests on the aggregated results from all models, datasets, and configurations. Since the Shapiro-Wilk test indicated that the data were not normally distributed ($p < 0.05$), we used the non-parametric Friedman test, which is suitable for comparing repeated measures across multiple conditions.

The Friedman test yielded a test statistic of 0.7118 and a $p$-value of 0.70. This result indicates that there are no statistically significant differences in model performance across the three repetition levels. The high $p$-value suggests that repeating questions 1, 3, or 5 times does not meaningfully affect LLM accuracy, supporting our conclusion that question repetition has no significant impact on performance. These findings align with the stability observed in our main results and reinforce the robustness of LLMs to variations in question repetition.

\end{document}